\documentclass[lettersize,journal]{IEEEtran}
\usepackage{algorithmic}
\usepackage{array}
\usepackage{textcomp}
\usepackage{stfloats}
\usepackage{verbatim}
\usepackage{algorithm,amsbsy,amsmath,amsfonts,amssymb,epsfig,bbm,mathrsfs,multirow,amsthm,xcolor,cite}
\usepackage{epstopdf}
\usepackage{graphicx,caption,subcaption}
\usepackage{setspace}
\usepackage[labelformat=simple]{subcaption}

\newcommand{\argmin}{\mathop{\rm argmin}}

\PassOptionsToPackage{hyphens}{url}
\usepackage{url}

\begin{document}

\title{Deep Transfer Learning: A Novel Collaborative Learning Model for Cyberattack Detection Systems in IoT Networks}

\author{Tran Viet Khoa, Dinh Thai Hoang, Nguyen Linh Trung, Cong T. Nguyen, \\Tran Thi Thuy Quynh, Diep N. Nguyen, Nguyen Viet Ha and Eryk Dutkiewicz

\thanks{T.~V.~Khoa is with the School of Electrical and Data Engineering, University of Technology Sydney, Sydney, NSW 2007, Australia and the Advanced Institute of Engineering and Technology (AVITECH), University of Engineering and Technology, Vietnam National University, Hanoi, Vietnam (e-mail: khoa.v.tran@student.uts.edu.au, khoatv.uet@vnu.edu.vn).}

\thanks{C.~T. Nguyen, D.~T.~Hoang, D.~N.~Nguyen, and E.~Dutkiewicz are with the School of Electrical and Data Engineering, University of Technology Sydney, Sydney, NSW 2007, Australia (e-mail: cong.nguyen@student.uts.edu.au, \{hoang.dinh, diep.nguyen, eryk.dutkiewicz\}@uts.edu.au).}

\thanks{N.~L.~Trung (corresponding author), T.~T.~T.~Quynh, and N.~V.~Ha are with the Advanced Institute of Engineering and Technology (AVITECH), University of Engineering and Technology, Vietnam National University, Hanoi, Vietnam (e-mail: \{linhtrung, quynhttt, hanv\}@vnu.edu.vn).}


}

\maketitle

\begin{abstract}
Federated Learning (FL) has recently become an effective approach for cyberattack detection systems, especially in Internet-of-Things (IoT) networks. By distributing the learning process across IoT gateways, FL can improve learning efficiency, reduce communication overheads and enhance privacy for cyberattack detection systems. However, one of the biggest challenges for deploying FL in IoT networks is the unavailability of labeled data and dissimilarity of data features for training. In this paper, we propose a novel collaborative learning framework that leverages Transfer Learning (TL) to overcome these challenges. Particularly, we develop a novel collaborative learning approach that enables a target network with unlabeled data to effectively and quickly learn ``knowledge'' from a source network that possesses abundant labeled data. It is important that the state-of-the-art studies require the participated datasets of networks to have the same features, thus limiting the efficiency, flexibility as well as scalability of intrusion detection systems. However, our proposed framework can address these problems by exchanging the learning ``knowledge'' among various deep learning models, even when their datasets have different features. Extensive experiments on recent real-world cybersecurity datasets show that the proposed framework can improve more than 40\% as compared to the state-of-the-art deep learning based approaches. 
\end{abstract}


\section{Introduction}\label{sec:Int}

\IEEEPARstart{I}{n} recent years, the rapid development of various technologies, such as 5G/6G, Industry 4.0, and Internet-of-Things (IoT), has enabled numerous applications to become an integral part in many aspects of our daily lives. However, such ever-fast growth has also led to an unprecedented massive amount of data and the proliferation of interconnected devices, e.g., sensors, smart cars, and cameras, which raises serious security and privacy concerns. Particularly, the increasing number of emerging applications has also brought forth many new types of cyberattacks. For example, the number of new (zero-day) cyberattacks has increased by 60\% from 2018 to 2019~\cite{purplesec_2021}. Besides the dire consequences to the economic, e.g., ransomware alone cost more than \$5 billion globally in 2017~\cite{ransom_2017}, cyberattacks pose serious threats to other areas with highly sensitive information such as healthcare and public security. As a result, cyberattack detection methods play a key role in detecting and promptly preventing consequences of cyberattacks in future IoT networks.

Recently, with outstanding classification ability, Machine Learning (ML) techniques, especially deep learning (DL), have been widely applied for cyberattack detection problems. Particularly, DL models can effectively learn the signatures of various cyberattack types. Moreover, DL models even can detect new types of attacks that have never been learned/trained before~\cite{khoa2022collaborative}. Nevertheless, DL-based cyberattack detection systems are also facing some practical challenges. Particularly, conventional DL approaches usually require a huge amount of data to achieve a high performance. However, in many applications, data are very difficult to collect because they are often stored locally on user devices such as IoT devices, smartphones, and wearable devices. This poses a threat to user privacy because sensitive data (e.g., location and private information) have to be sent over the network and stored at the centralized server for processing. Besides the privacy concerns, transmitting such a collectively large amount of data also imposes an extra communication burden over the network. Consequently, these limitations have been hindering the effectiveness of DL techniques in cyberattack detection~systems.

To address these problems, Federated Learning (FL) has emerged to be a highly effective solution. Unlike conventional DL techniques that collect data and train the global model at a central server, FL enables the learning process to be distributed across all devices. Particularly, instead of sending data to a central server, the local data can be used to train a global model locally on each user device. Then, the obtained model weights of each device are periodically sent to a central server for aggregation. Afterward, the aggregated weights are sent back to all devices to update their local models' weights. Since only the weights are transmitted in FL, both the privacy and communication overhead issues can be mitigated~\cite{Hoang_federatedsurvey}.   

Despite its effectiveness, FL is still facing some challenges. Particularly, FL only performs well if the training data and the predicting data are independent and identically distributed (i.i.d). Consequently, they are not robust to the changes in the system, e.g., changes in network traffic due to the mobility of users, new types of devices participating in the network, and so on. Moreover, the performance of FL largely relies on the availability of labeled data. However, acquiring sufficient labeled data might be costly and time-consuming. Even if the data are available, the participated user data usually have different structures such as features. This leads to difficulty or even mistakes when FL aggregates the global model. Consequently, they may not be suitable for the intensive training process of FL~\cite{Hoang_federatedsurvey}~\cite{aledhari2020federated}.

To address these limitations, transfer learning (TL) has been emerging as a promising solution, especially for problems related to heterogeneous training data~\cite{nguyen2021transfer,niu2020decade,zhuang2020comprehensive}. Unlike DL and FL techniques that are trained only for a specific problem, TL can utilize ``knowledge'' from rich resource data to enhance the training process and performance of the ML models. Particularly, by transferring ``knowledge'' from similar scenarios with a lot of high-quality data, TL can address the lack of labeled data for the target networks. Moreover, the TL can exchange ``knowledge'' even if the data features of the target and source networks are not very similar~\cite{nguyen2021transfer,xu2022secure}. However, if the data features are too different, TL might even make the learning process worse than that without using TL, i.e., negative transfer~\cite{nguyen2021transfer,niu2020decade,zhuang2020comprehensive}. In the context of cyberattack detection for IoT networks, negative transfer might be a serious problem since different networks may have various types of devices generating different data.

In this paper, we propose a novel collaborative learning framework that utilizes the strengths of both TL and FL to address the limitations of conventional DL-based cyberattack detection systems. Particularly, we consider a scenario with two different IoT networks\footnote{The cases with multiple networks can be straightforwardly extended, e.g., by scheduling for networks to exchange information in order.}. The first network (source network) has an abundant labeled data resource, while the second network (source network) has very little data resource (and most of them are unlabeled). Here, unlike most of the current works that assume that the data at these networks have the same features~\cite{ferrag2021federated}, we consider a much more practical and general case in which data at these two networks may have different features. To address the problem of dissimilar feature spaces of the target and source networks, we propose to transform them into a new joint feature-space. In this case, at each learning round of the federated learning process, trained models of target and source networks can be exchanged through the joint feature-space. Thus, by periodically exchanging and updating the trained model, the target network can eventually achieve the converged trained deep neural network that can predict attacks with high accuracy (thanks to useful ``knowledge'' transferred from the source network). Besides the exchanging and updating the learning model iteratively, we use a small number of mutual samples between two networks to mitigate the negative transfer learning. More importantly, unlike FL where networks try to train a joint global model, our proposed framework enables the participating networks to obtain their particular trained models that are specific to their networks, i.e., better predict attacks for particular networks with different data structures. Extensive experiments on recent real-world datasets, including N-BaIoT~\cite{autoencoder1}~\cite{autoencoder2}, KDD~\cite{kdd}, NSL-KDD~\cite{nslkdd} and UNSW~\cite{unsw} show that our proposed framework can achieve an accuracy of up to 99\% and an improvement of up to 40\% over the unsupervised learning approach. The main contributions of this paper can be summarized as follows:

\begin{itemize} 
	\item We propose a novel collaborative learning framework that can effectively detect cyberattacks in decentralized IoT systems. By combining the strengths of FL and TL, our proposed framework can improve learning efficiency and the accuracy of cyberattack detection in comparison with the conventional DL-based cyberattack detection systems.
	
	\item We propose an effective transfer learning approach that can allow the deep learning model from the rich-data network to transfer useful knowledge to the low-data network even they have different features for cyberattack detection in IoT networks.
		
	\item We perform extensive experiments on recent real-world datasets including N-BaIoT, KDD, NSL-KDD, and UNSW to evaluate the performance of the proposed collaborative learning framework. The results show that our proposed approach can achieve an accuracy of up to 99\% and an improvement of up to 40\% over the unsupervised learning approach.  
\end{itemize} 

The rest of this paper is organized as follows. We first discuss related works in Section~\ref{sec:Relatedwork}. We then propose the federated transfer learning model for cyberattack detection in Section~\ref{sec:Sys}. After that, simulation settings and results are discussed in Section~\ref{sec:setting}. Finally, we conclude the paper in Section~\ref{sec:Conc}.

\section{Related work}\label{sec:Relatedwork}
\subsection{Deep Learning for Cyberattack Detection}
There have been a rich literature proposing DL approaches for cyberattack detection. In~\cite{vinayakumar2019deep}, a deep neural network (DNN) model is developed to detect zero-day attacks based on two types of data, i.e., network activities and local system activities. The results show that for most of the datasets, the proposed DNN can achieve a higher detection accuracy and lower false-positive rate compared to those of the other conventional machine learning classifiers such as K-Nearest Neighbors (KNN), and Support Vector Machine (SVM). Another DL approach is proposed in~\cite{Khoi} to detect cyberattacks in the mobile cloud computing environments. The main difference between~\cite{vinayakumar2019deep} and~\cite{Khoi} is that the approach in~\cite{Khoi} consists of a feature analysis phase before the learning phase. In the analysis phase, the datasets are analyzed to identify meaningful features, thereby reducing the data dimension and computational complexity. Experiments on the KDD~\cite{kdd}, NSL-KDD~\cite{nslkdd}, and the UNSW~\cite{unsw} datasets show that the proposed approach can achieve a detection accuracy of up~to~97.1\%. 

\subsection{Federated Learning for Cyberattack Detection}
With the advent of FL, the research focus has recently shifted towards applying this framework for cyberattack detection, especially in environments with numerous devices such as IoT and mobile edge networks. In~\cite{abeshu2018deep}, an FL framework is proposed for cyberattack detection in an edge network. In this network, the data for intrusion detection are stored locally at each edge node. The edge nodes train their data locally and send their models' weights to an FL server for aggregating. After aggregation, the FL server sends the weights back to all edge nodes. In this way, each edge node can benefit from the other nodes' data and training while protecting its privacy and reducing the network's communication burden. Experiments with the NSL-KDD datasets show that the proposed approach can achieve an accuracy of up to 99.2\%. Another FL approach is proposed in~\cite{li2020deepfed} for attack detection in industrial cyber-physical systems. In the considered setting, there are multiple cyber-physical systems acting as FL nodes. However, unlike the previous frameworks, the authors propose a novel architecture combining a convolution neural network (CNN) and a gated recurrent unit for training at each FL node. Experiments with self-collected data show that the proposed approach can outperform other state-of-the-art approaches, e.g.,~\cite{nguyen2019diot,schneble2019attack,chen2020fedhealth}, with an accuracy up to 99.2\%. However, because of the limitations of FL as presented in the previous section, the learning model can only combine data with the same features and labels.

\subsection{Transfer Learning for Cyberattack Detection}
Although FL techniques can effectively address the privacy and communication load concerns of conventional ML for cyberattack detection, they are still facing some challenges. Particularly, FL approaches usually require high-quality and labeled data for training. However, collecting and labeling such data is expensive and time-consuming, especially for large-scale systems. On the other hand, unlabeled data are often abundant in environments such as IoT and mobile edge networks. Thus, a deep TL approach is proposed for IoT intrusion detection in~\cite{vu_deep_2020} based on network activities, which can utilize both labeled and unlabeled data. In this approach, the authors employ two AEs. The first AE is trained with labeled data, while the second AE is trained with unlabeled data. Then, the knowledge is transferred from the first AE to the second AE by minimizing the Maximum Mean Discrepancy (MMD) distances between their weights. Experiments over nine IoT datasets were conducted to show that the proposed approach can achieve higher Area Under the Curve (AUC) scores compared to those of several other approaches. 

Besides analyzing network traffic, another approach to detect cyberattacks is to analyze the devices' fingerprints. Particularly, attackers may try to impersonate a device in the system by copying its signal. For this kind of attack, ML techniques can be used to detect if the signals are coming from the real device or the malicious device. TL approaches such as~\cite{sharaf_transfer_2015,zhao_the_2018,sharaf_on_2016,dabbagh_authentication_2019} are proposed to identify cyberattacks based on device fingerprints. Among them,~\cite{sharaf_on_2016} and~\cite{dabbagh_authentication_2019} leverage the environmental effects to classify signals from devices. To improve the classification accuracy and address the lack of data, these approaches transfer the knowledge from nearby devices (since they share similar environmental effects). On the other hand,~\cite{sharaf_transfer_2015} and~\cite{zhao_the_2018} leverage the knowledge from previous experiences, i.e., data collected in the past. These past data are then combined with the current data for training, thereby addressing the lack of fingerprint data. 

Unlike all the abovementioned approaches, the collaborative learning framework proposed in this paper can leverage the strengths of both FL and TL to address limitations of ML-based intrusion detection systems, e.g., lack of labeled data, privacy and heterogeneous data feature space. Moreover, in our approach, each IoT network has a separated model that is fine-tuned specifically for that network, therefore the model is more effective for that network's cyberattack detection compared to FL frameworks with a single model for all networks. Furthermore, our proposed system model can utilize knowledge from both source and target data in the network instead of only transferring knowledge from a single source as proposed in most of the mentioned TL frameworks~\cite{vu_deep_2020,wen_time_2019,zhao_the_2018,sharaf_transfer_2015}, thereby mitigating the negative transfer problem. 

\begin{figure*}[t!]
	\centering
	\includegraphics[width=0.76\linewidth]{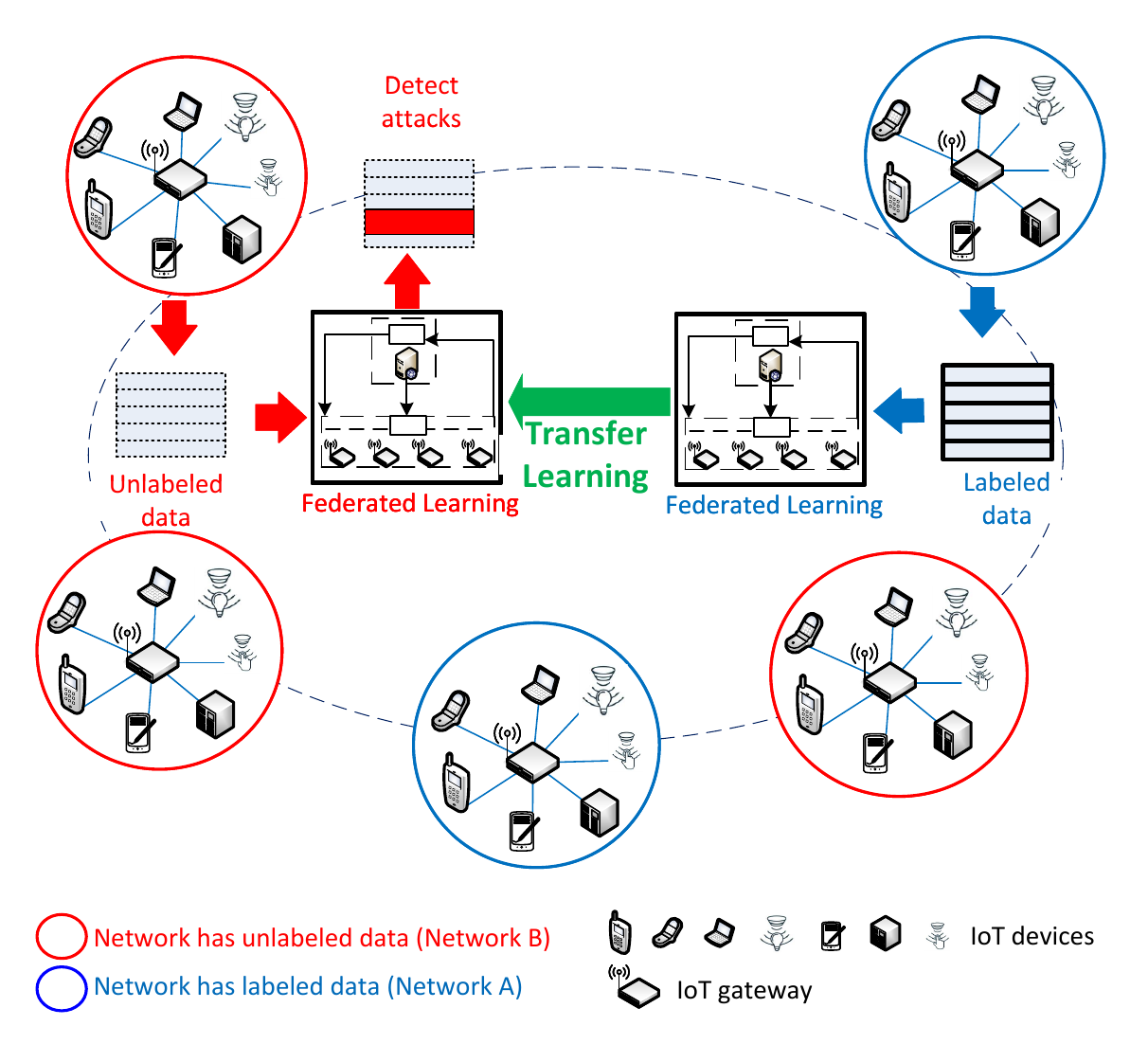}
	\caption{Illustration of a system model for cyber attack detection in IoT networks.}
	\label{fig:systemmodel}
\end{figure*} 

\section{Proposed Federated Transfer Learning Framework for Cyberattack Detection in IoT Networks}\label{sec:Sys}

\subsection{System Model}

The conventional FL model requires to use a centralized server to maintain and aggregate all the trained models in the whole learning process. However, this may lead to a high cost to maintain and may not be effective to deploy in IoT networks. Thus, in this work, we propose a federated transfer learning model that allows the learning process to be performed more flexibly and effectively in IoT environments. In particular, we consider a network which has unlabeled data (e.g., Network B as illustrated in Fig.~\ref{fig:systemmodel}), and it wants to learn more knowledge from other networks with abundant labeled data. In this case, this network will connect with a target network (e.g., Network A as illustrated in Fig.~\ref{fig:systemmodel}) and nominate itself as a centralized node which can train its own data as well as perform transfer learning to exchange knowledge with the target network. 

We denote a labeled cybersecurity dataset $D_A= $ $\{X^A, Y^A, F^A\}$ of Network A with $(X^A, Y^A) =$ $ \{x^A_1, y^A_1, x^A_2, y^A_2, \ldots, x^A_{M_A}, y^A_{M_A}\}$ where $M_A$ is the number of samples of dataset A. In contrast, Network B has an unlabeled cybersecurity dataset $D_B=\{X^B,F^B\}$ with $(X^B)=\{x^B_1,x^B_2, \ldots,x^B_{M_B}\}$ where $M_B$ is the number of samples of dataset B. $F^A,F^B$ are the feature spaces of Network A and Network B, respectively. The proposed model will perform transfer learning between two neural network by minimizing the total loss $J$ to predict the label $P(z^B)$ for the unlabeled dataset of Network B. In this way, the network can help to improve the accuracy in identifying network traffics by learning useful knowledge from other labeled networks. Each network can be managed by an IoT gateway and possesses its own private dataset. The IoT gateway uses its deep learning model to detect normal and abnormal traffics. It is important to note that, unlike conventional FL approaches~\cite{Khoa_WCNC2020}, in this work, we consider a practical scenario in which the datasets of networks may have different features.

\subsection{Proposed Federated Transfer Learning Approach for Cyberattack Detection}
\begin{table}[!b]
	\centering
	\caption{Notations.}
	\label{Table:1}	
	\begin{tabular}{|l|l|l}
	\cline{1-2}
	\multicolumn{1}{|l|}{\textbf{Notation}} & \textbf{Description}                                    &  \\ \cline{1-2}
	$X$                                     & The total samples of a dataset                          &  \\ \cline{1-2}
	$Y$                                     & The labels of a dataset                                 &  \\ \cline{1-2}
	$F$                                     & The feature space of a dataset                                   &  \\ \cline{1-2}
	$x$                                     & A dataset sample                                        &  \\ \cline{1-2}
	$D$                                     & Network                                                 &  \\ \cline{1-2}
	$M_A, M_B$                              & The number of samples of dataset A, B, \\ & respectively      &  \\ \cline{1-2}
	$M_C$                                   & The number of predicted labels                          &  \\ \cline{1-2}
	$M_{AB}$                                & The overlapping samples between dataset A\\ & and dataset B &  \\ \cline{1-2}
	$W_A, W_B$                              & The parameter matrices of models A and B, \\ & respectively       &  \\ \cline{1-2}
	$Z_A, Z_B$                              & The outputs of models A and B, \\ & respectively                  &  \\ \cline{1-2}
	$z$                                     & The output of an input sample after \\ & learning model   &  \\ \cline{1-2}
	$j$                                     & The loss of an input sample                             &  \\ \cline{1-2}
	$J$                                     & The loss function                                       &  \\ \cline{1-2}
	$\gamma, \lambda$                       & Weight parameters                                       &  \\ \cline{1-2}
	$w$                                     & Training parameters                                                     &  \\ \cline{1-2}
\end{tabular}
\end{table}

\begin{algorithm}[t]
	\algsetup{linenosize=\tiny}
	\caption{Federated Transfer Learning Algorithm: Training Process}
	\label{al:FTL_training}
	\begin{algorithmic}[1]
		\STATE \textbf{Input:} The learning rate $\eta$, the weight parameter $\gamma,\lambda$, the maximum iteration $T$, the tolerance $t$ and Network A and Network B initialize model parameters $W^A,W^B$;
		\STATE \textbf{Output:} The trained model parameter $W^A,W^B$;
		\STATE $iteration=0$
		\WHILE {$ iteration \leq T$}
		\STATE Network A performs:
		\STATE $z^A_i = h^A_i * x^A_i$ for $i \in D_A; $
		\STATE Send $\{z^A_i,y^A_i\}$ to Network B;
		\STATE Network B performs:
		\STATE $z^B_i = h^B_i * x^B_i$ for $i \in D_B; $
		\STATE Send $\{z^B_i\}$ to Network A;
		\STATE Network A performs:
		\STATE Compute $\frac{\partial J}{\partial w^A_i}$ and $J^A$, then send them to Network B;
		\STATE Network B performs:
		\STATE Compute $\frac{\partial J}{\partial w^B_i}$, $J^B$ and $J^{AB}$, then send them to Network A;	
		\STATE Network A performs:
		\STATE Update $w^A_l=w^A_l - \eta\frac{\partial J}{\partial w^A_i};$
		\STATE Network B performs:
		\STATE Update $w^B_l=w^B_l - \eta\frac{\partial J}{\partial w^B_i};$
		\IF {$J_{prev} - J \leq t$}
			\STATE Send stop signal to Network B;
			\STATE Break.
		\ELSE
			\STATE $J_{prev} = J$;
			\STATE $iteration = iteration + 1$;
			\STATE continue;
		\ENDIF
		\ENDWHILE
	\end{algorithmic}
\end{algorithm}

\begin{figure}[ht]
	\begin{subfigure}{\linewidth}
		\centering
		\includegraphics[width=\linewidth]{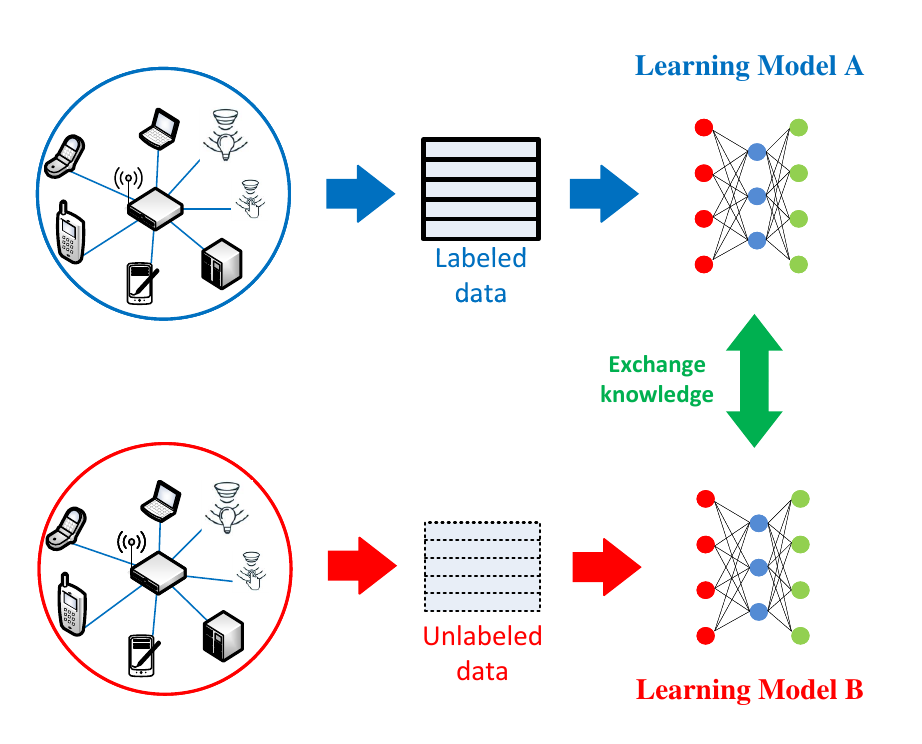}  
		\caption{FTL training process.}
		\label{fig:FTL_train}
	\end{subfigure}
	\begin{subfigure}{\linewidth}
		\centering
		\includegraphics[width=\linewidth]{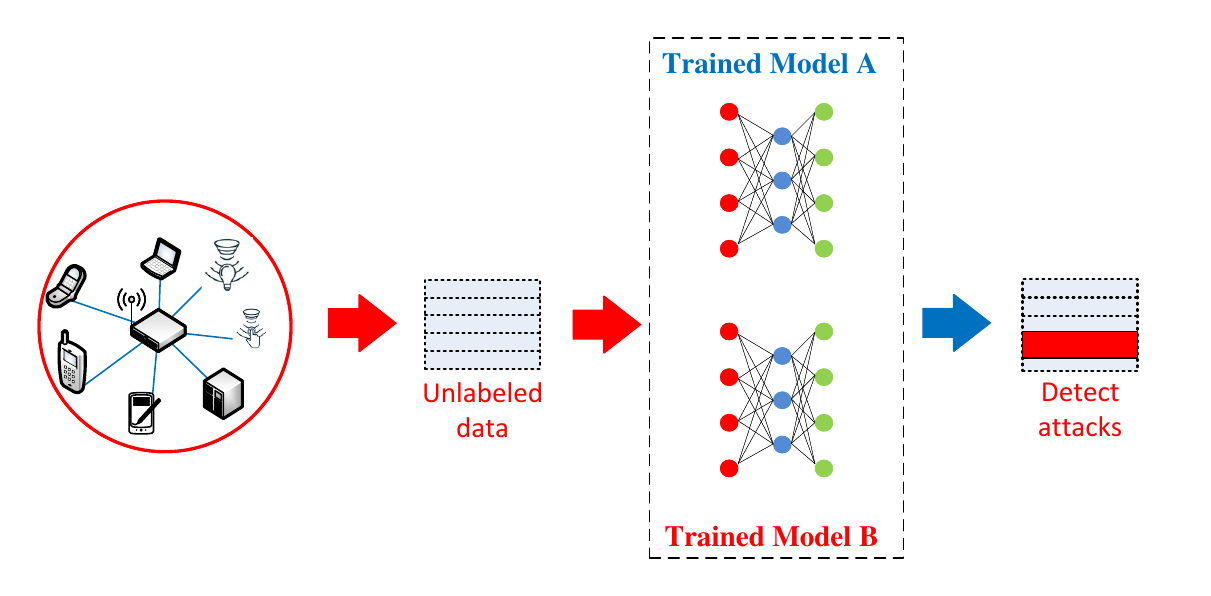}  
		\caption{FTL predicting process.}
		\label{fig:FTL_predict}
	\end{subfigure}
	\caption{The FTL algorithm.}
	\label{fig:FTL}
\end{figure}

In this section, we propose a highly-effective federated transfer learning model that can exchange knowledge between an unlabeled network and multiple networks which may have different features. To better analyze the impact of our proposed approach, we consider a specific scenario in which one labeled network is used as a source network to support an unlabeled network (i.e., target network). The scenario with one unlabeled network and multiple labeled networks can be straightforwardly extended, and we leave it for future study. Fig.~\ref{fig:FTL} describes the training and predicting processes of FTL algorithm that we use in this case. The table of notations is presented in Table~\ref{Table:1}. As described in previous section, Network A, Network B have their dataset $D^A, D^B$, respectively. They also have their model parameters called $W^A$ and $W^B$. The outputs of two neural networks are calculated as follows:
\begin{subequations}
	\label{func1}
    \begin{align}
    	Z^A &= W^A*X^A, \\
    	Z^B &= W^B*X^B.
    \end{align} 
\end{subequations}

We need to find the prediction function $P(z^B_j)=P(z^A_1, y^A_1,\ldots,z^A_{M_A}, y^A_{M_A}, z^B_j)$ to predict the output of Network B. To find a high-quality predict function, we first need to minimize the loss function using the labeled dataset as follows:

\begin{equation}
\arg\min_{W^A,W^B} J^B = \sum_{i}^{M_c} j^B(y^A_i,P(z^B_i)),
\label{func4}
\end{equation}
where $M_c$ is the number of predicted labels, and $j^B$ represents the loss of the loss function which depends on the type of output or mechanism, i.e., the logistic loss function~\cite{FTL2} with the predicted value $\mathbf{z}$ and the labeled~$\mathbf{y}$: 
\begin{equation}
	j^B(\mathbf{z,y})=\log\big(1 + \exp(-\mathbf{z} \times \mathbf{y})\big).
	\label{func2}
\end{equation} 
In addition, datasets A and B may have some overlapping samples, and thus we can use these samples to optimize the loss function. We denote $M_{AB}$ as the overlapping samples between dataset A and dataset B. We need to minimize the alignment loss function between A and B as follows:

\begin{equation}
\argmin_{W^A,W^B} J^{AB} = -\sum_{i}^{M_{AB}} j^{AB}(z^A_i,z^B_i),
\label{func5}
\end{equation}
where $j^{AB}$ represents the alignment loss function. The common alignment loss function can be represented in modulus $j^{AB}=||z^A_i-z^B_i||^2$ or angle $j^{AB}=-z^A_i*z^B_i$. Lastly, we add the regularization $J^A_R=\sum_{l}^{L_A}{||w^A_l||^2}$ and $J^B_R=\sum_{l}^{L_B}{||w^B_l||^2}$ in which $L_A$ and $L_B$ are the numbers of layers in neutron Network A and Network B, respectively, to find the final loss function that needs to be minimized: 

\begin{equation}
\argmin_{W^A,W^B} J = J^B + \gamma J^{AB} + \frac{\lambda}{2} (J^A_R + J^B_R),
\label{funcJ}
\end{equation}
where $\gamma$ and $\lambda$ are the weight parameters. The gradient for updating $W^A,W^B$ are calculated by the following formula:

\begin{equation}
\frac{\partial J}{\partial w^i_l} = \frac{\partial J^B}{\partial w^i_l} + \gamma \frac{\partial J^{AB}}{\partial w^i_l} + \lambda w^i_l.
\label{funcgradient}
\end{equation}

The training process is presented in Algorithm~\ref{al:FTL_training}. Specifically, we first initialize $W^A$ and $W^B$. Next, we calculate $z^A_i$ and $z^B_i$ from the input samples of dataset A ($D^A$) and dataset B ($D^B$) as shown in Equation~(\ref{func1}). Then, Network A sends $\{z^A_i,y^A_i\}$ to Network B to calculate $J^B$, the alignment loss function $J^{AB}$ and the gradients of $J^B$ as shown in Equations~(\ref{func4}),~(\ref{func5}),~(\ref{funcJ}) and~(\ref{funcgradient}), respectively. Similarly, Network B sends $\{z^B_i\}$ to Network A to calculate $J^A$ as in Equation~(\ref{funcJ}). In Equation~(\ref{func5}), we use $M_{AB}$ as the mutual samples of two datasets. For example, the same IoT devices are attacked by the same types of cyberattacks in different networks. Each network extracts the attack data with different features, e.g., Network A uses timeslot, packet header, ip address while Network B uses MAC address, error packets, frame header. The number of mutual samples is an important factor that strongly supports the learning process between two networks (we will explain it more details in Section~\ref{sec:setting}). After that, we calculate the final loss function $J$ and the gradient as in Equation~(\ref{funcJ}) and Equation~(\ref{funcgradient}). Finally, Network A and Network B update their model parameters based on the gradient and loss functions. This process continuously repeats until the system converges or reaches the maximum number of iterations to minimize the final loss function in Equation~(\ref{funcJ}).

When the training completes, the prediction process described in the Algorithm~\ref{al:FTL_predicting} is called to predict the final result of the unlabeled dataset $D_B$. In this process, both Network A and Network B have their trained models. Similar to the training process, the dataset $D_B$ firstly goes through the the trained model of Network B to calculate $Z^B$. Then, Network B sends $Z^B$ to Network A to archive the transfer learning knowledge from trained model of Network A. Network A predicts the results and sends them back to Network B to classify the attack and normal behaviors of the network.

\begin{algorithm}[t]
	\algsetup{linenosize=\tiny}
	\caption{Federated Transfer Learning Algorithm: Predicting Process}
	\label{al:FTL_predicting}
	\begin{algorithmic}[1]
		\STATE \textbf{Input:} The model parameters $W^A,W^A$ and dataset $X_B$;
		\STATE \textbf{Output:} The prediction $Y^B$;
		\STATE Network B performs:
		\STATE $z^B_i = h^B_i * x^B_i$ for $i \in D_B; $
		\STATE Send $\{z^B_i\}$ to Network A;
		\STATE Network A performs:
		\STATE Compute $P(z^B_i)=W^A[z^B_i]$ and send it to Network B.	
	\end{algorithmic}
\end{algorithm}

\subsection{Evaluation Methods}

As mentioned in~\cite{confusion_matrix1,confusion_matrix2}, the confusion matrix is typically used to evaluate system performance, especially for intrusion detection systems. We denote TP, TN, FP, and FN to be ``True Positive'', ``True Negative'', ``False Positive'', and ``False Negative'', respectively. The Receiver Operator Characteristic (ROC) is created by plotting the TPR over FPR at different thresholds. Then, we use Area Under the Curve (AUC) to evaluate the performance of the algorithm in the following formula: 
\begin{equation}
\xi=\int_{x=0}^{1} \mbox{TP}(\mbox{FP}^{-1}(x)) \,dx.
\end{equation}

In our experiments, we randomly select samples from original dataset to test the algorithm. In this scenario, the $p$-value is often used to evaluate the results of random tests, and is given by
\begin{equation}
p = F(\xi|\mu,\sigma)= \frac{1}{\sigma \sqrt{2\pi}} \int_{-\infty}^\xi \mathrm{e}^{\frac{-(t-\mu)^2}{2\sigma^2}}\,\mathrm{d}\xi,
\end{equation}  
in which $\mu$ is the mean and $\sigma$ is the standard deviation. The results are calculated by the significant number with the following formula:
\begin{equation}
Sig = F^{-1}(p|\mu,\sigma) = \{\xi:F(\xi|\mu,\sigma)=p\}, 
\label{equal:sig}
\end{equation}
where $Sig$ is the significant number that represents the results of 30 random runs and the confidence of this number is calculated by $conf = 1-p$. In a normal situation, $p$ is considered confidence when it has values around 0.01 and 0.05, corresponding to the confidence of significant numbers is around 99\% and 95\%.

\section{Performance Analysis}\label{sec:setting}

\subsection{Datasets}

In this experiment, we use four popular cybersecurity datasets namely the N-BaIoT~\cite{autoencoder1}~\cite{autoencoder2}, KDD~\cite{kdd}, NSL-KDD~\cite{nslkdd} and UNSW~\cite{unsw} datasets to evaluate the performance of the proposed method. The Network-based Detection of IoT Botnet Attacks (N-BaIoT) dataset~\cite{autoencoder1}~\cite{autoencoder2} includes the information collected in the setup network about the normal and attack situation. The attack was performed by servers to nine IoT devices and the total network behavior was captured by the sniffer server to extract dataset. This dataset is characterized by 115 features for both normal and attack behaviors. In this dataset, the attack type is the Distributed Denial of Service (DDoS) which was implemented by two well-known botnets, namely Mirai and BASHLITE. The BASHLITE botnet includes 5 types of attacks, i.e., network scanning (scan), spam data sending (junk), UDP flooding (udp), TCP flooding (tcp), and the join of sending spam data and opening port to specific IP address (combo). Besides BASHLITE, the Mirai botnet also includes 5 types of attacks, i.e., scan, ACK flooding (ack), SYN flooding (syn), udp, and optimized UDP flooding (udpplain). 

In addition to IoT datasets, we also want to evaluate our proposed solution on some classical intrusion detection datasets, i.e., KDD~\cite{kdd}, NSL-KDD~\cite{nslkdd} and UNSW~\cite{unsw} datasets. The KDD dataset~\cite{kdd} includes many different kinds of network attacks simulated in military network environment. The KDD dataset has 41 features and it classifies attacks into 4 groups including Denial of Service (DoS), Probe, User to Root (U2R), Remote to Local (R2L). The NSL-KDD dataset~\cite{nslkdd} inherits the properties from KDD~\cite{kdd} dataset such as the features and types of attacks but eliminates the redundant samples in the training dataset and the duplicated samples in the testing dataset. Although both KDD and NSL-KDD datasets are well-known and used in many research works, they were developed long time ago. Thus, some modern attacks were not involved. Therefore, a recent dataset, i.e., UNSW dataset~\cite{unsw}, is considered in this work. Unlike KDD and NSL-KDD, the feature space of this dataset includes 42 types and 9 kinds of attacks, namely DoS, Backdoors, Worms, Fuzzers, Analysis, Reconnaissance, Exploits, Shellcode, and Generic.

\begin{figure}[t]
	\centering     
	\includegraphics[width=.95\linewidth]{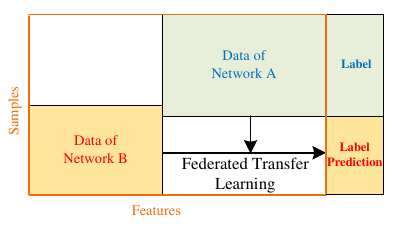}
	\caption{The data of participated networks used in this experiment.}
	\label{fig:Dataset_used}
\end{figure}

\begin{table*}
	\centering
	\begin{subtable}{0.3\linewidth}\centering
		\begin{tabular}{|l|l|l|}
			\hline
			& FTL    & UDL \\ \hline
			IoT1   & \textbf{85.771} & 45.753     \\ \hline
			IoT2   & \textbf{83.795} & 63.171     \\ \hline
			IoT3   & \textbf{94.286} & 80.453     \\ \hline
			IoT4   & \textbf{79.241} & 77.885     \\ \hline
			IoT5   & \textbf{90.605} & 81.876     \\ \hline
			IoT6   & \textbf{91.179} & 82.703     \\ \hline
			IoT7   & \textbf{90.670} & 85.183     \\ \hline
			IoT8   & \textbf{82.960} & 65.256     \\ \hline
			IoT9   & \textbf{83.222} & 73.072     \\ \hline
			KDD    & \textbf{99.315} & 80.477     \\ \hline
			NSLKDD & \textbf{98.485} & 83.025     \\ \hline
			UNSW   & \textbf{97.072} & 68.449     \\ \hline
		\end{tabular}
		\caption{The results with $p=1$. }
		\label{table:results_p=1}
	\end{subtable}
	\hfill
	\begin{subtable}{0.3\linewidth}\centering
		\begin{tabular}{|l|l|l|}
			\hline
			& FTL    & UDL \\ \hline
			IoT1   & \textbf{87.398} & 49.770     \\ \hline
			IoT2   & \textbf{85.672} & 65.793     \\ \hline
			IoT3   & \textbf{94.896} & 81.070     \\ \hline
			IoT4   & \textbf{81.672} & 77.885     \\ \hline
			IoT5   & \textbf{91.517} & 82.013     \\ \hline
			IoT6   & \textbf{92.059} & 82.703     \\ \hline
			IoT7   & \textbf{92.030} & 86.013     \\ \hline
			IoT8   & \textbf{85.197} & 68.161     \\ \hline
			IoT9   & \textbf{85.072} & 73.078     \\ \hline
			KDD    & \textbf{99.395} & 81.304     \\ \hline
			NSLKDD & \textbf{98.534} & 83.450     \\ \hline
			UNSW   & \textbf{97.141} & 69.124     \\ \hline
		\end{tabular}
		\caption{The results with $p=3$. }
		\label{table:results_p=3}
	\end{subtable}
	\hfill
	\begin{subtable}{0.3\linewidth}\centering
		\begin{tabular}{|l|l|l|}
			\hline
			& FTL    & UDL \\ \hline
			IoT1   & \textbf{88.259} & 51.897     \\ \hline
			IoT2   & \textbf{86.666} & 67.181     \\ \hline
			IoT3   & \textbf{95.220} & 81.397     \\ \hline
			IoT4   & \textbf{82.959} & 77.885     \\ \hline
			IoT5   & \textbf{92.000} & 82.085     \\ \hline
			IoT6   & \textbf{92.525} & 82.703     \\ \hline
			IoT7   & \textbf{92.750} & 86.453     \\ \hline
			IoT8   & \textbf{86.381} & 69.700     \\ \hline
			IoT9   & \textbf{86.052} & 73.082     \\ \hline
			KDD    & \textbf{99.438} & 81.742     \\ \hline
			NSLKDD & \textbf{98.561} & 83.675     \\ \hline
			UNSW   & \textbf{97.177} & 69.482     \\ \hline
		\end{tabular}
		\caption{The results with $p=5$. }
		\label{table:results_p=5_10000}
	\end{subtable}
	\caption{The results with multiple datasets in CASE 1.}
	\label{table:results_10000}
\end{table*}

\begin{table*}[!b]
	\centering
	\resizebox{0.7\linewidth}{!}{%
		\begin{tabular}{|c|c|c|c|c|}
			\hline
			Dataset & Device name                                   & Features of & Features of & Total \\ &&Network A &Network B& features\\ \hline
			IoT1    & Danmini\_Doorbell                             & 85 & 30 & 115            \\ \hline
			IoT2    & Ecobee\_Thermostat                            & 85 & 30 & 115            \\ \hline
			IoT3    & Ennio\_Doorbell                               & 85 & 30 & 115            \\ \hline
			IoT4    & Philips\_B120N10\_Baby\_Monitor               & 85 & 30 & 115            \\ \hline
			IoT5    & Provision\_PT\_737E\_Security\_Camera         & 85 & 30 & 115            \\ \hline
			IoT6    & Provision\_PT\_838\_Security\_Camera          & 85 & 30 & 115            \\ \hline
			IoT7    & Samsung\_SNH\_1011\_N\_Webcam                 & 85 & 30 & 115            \\ \hline
			IoT8    & SimpleHome\_XCS7\_1002\_WHT\_Security\_Camera & 85 & 30 & 115            \\ \hline
			IoT9    & SimpleHome\_XCS7\_1003\_WHT\_Security\_Camera & 85 & 30 & 115            \\ \hline
			KDD     & - 											& 31 & 10 & 41             \\ \hline
			NSLKDD  & - 											& 31 & 10 & 41             \\ \hline
			UNSW    & - 											& 31 & 11 & 42             \\ \hline
		\end{tabular}%
	}
	\caption{Dataset preparation}
	\label{table:dataset}
\end{table*}

\subsection{Experiment Setup}

In this section, we carry out experiments using all the aforementioned datasets to evaluate the performance of the proposed solution. In this experiment, we denote IoT1-9 as the dataset names of nine IoT devices. Table~\ref{table:dataset} describes the total features and the representative names of datasets that we use in this experiment. Fig.~\ref{fig:Dataset_used} also describes the separated data in each dataset in this experiment. In this experiment, the participated data are randomly selected from the dataset. Then, the selected data are separated into label data (data of Network A) and unlabeled data (data of Network B) with different features as described in Table~\ref{table:dataset}. These data have about 10\% mutual samples of total dataset samples. We experiment with two cases, i.e., the first one is with 2000 unlabeled data and 9577 labeled data (CASE 1), the second one is with 10000 unlabeled data and 47893 labeled data (CASE 2). 

In this setup, we consider a baseline solution with the state-of-the-art unsupervised deep learning model (UDL) which clusters the unlabeled data into normal and attack behaviors based on autoencoder and k-means techniques~\cite{goodfellow}. The unsupervised deep learning model includes an autoencoder and k-nearest neighbor to cluster the unlabeled data. In addition, we consider the second baseline solution that uses both supervised and unsupervised datasets to feed the FTL learning models. The FTL will exchange the knowledge from the supervised learning model and the unsupervised learning model to improve the accuracy of learning as well as increase the precise of identifying attack and normal behaviors of the unlabeled data. Then, we measure the AUC of this process 30 times to calculate the signification number of the AUC series results with both baseline solutions. Finally, we plot the reconstruction errors to analyze the convergence of the FTL algorithm for all datasets.  

\subsection{Experimental Results}\label{subsec:result}

In this section, we show the results of our experiments with different kinds of cybersecurity datasets.

\subsubsection{Accuracy Comparison}

\begin{table*}
	\centering
	\begin{subtable}{0.3\linewidth}\centering
		\begin{tabular}{|l|l|l|}
			\hline
			& FTL    & UDL \\ \hline
			IoT1   & \textbf{90.371} & 49.783 \\ \hline
			IoT2   & \textbf{68.193} & 62.591 \\ \hline
			IoT3   & \textbf{94.525} & 83.411 \\ \hline
			IoT4   & \textbf{87.050} & 77.725 \\ \hline
			IoT5   & \textbf{86.535} & 81.954 \\ \hline
			IoT6   & \textbf{87.214} & 82.555 \\ \hline
			IoT7   & \textbf{97.662} & 79.517 \\ \hline
			IoT8   & \textbf{84.609} & 52.702 \\ \hline
			IoT9   & \textbf{90.095} & 63.803 \\ \hline
			KDD    & \textbf{99.535} & 84.333 \\ \hline
			NSLKDD & \textbf{98.858} & 81.164 \\ \hline
			UNSW   & \textbf{97.049} & 66.329 \\ \hline
		\end{tabular}
		\caption{The results with $p=1$. }
		\label{table:results_p1_5000}
	\end{subtable}
	\hfill
	\begin{subtable}{0.3\linewidth}\centering
		\begin{tabular}{|l|l|l|}
			\hline
			& FTL    & UDL \\ \hline
			IoT1   & \textbf{91.497} & 54.079 \\ \hline
			IoT2   & \textbf{72.573} & 65.681 \\ \hline
			IoT3   & \textbf{95.073} & 83.565 \\ \hline
			IoT4   & \textbf{88.538} & 77.781 \\ \hline
			IoT5   & \textbf{88.150} & 82.160 \\ \hline
			IoT6   & \textbf{88.638} & 82.664 \\ \hline
			IoT7   & \textbf{97.928} & 81.400 \\ \hline
			IoT8   & \textbf{86.691} & 57.318 \\ \hline
			IoT9   & \textbf{90.959} & 65.559 \\ \hline
			KDD    & \textbf{99.562} & 84.423 \\ \hline
			NSLKDD & \textbf{98.885} & 81.976 \\ \hline
			UNSW   & \textbf{97.121} & 66.901 \\ \hline
		\end{tabular}
		\caption{The results with $p=3$.}
		\label{table:results_p3_50000}
	\end{subtable}
	\hfill
	\begin{subtable}{0.3\linewidth}\centering
		\begin{tabular}{|l|l|l|}
			\hline
			& FTL    & UDL \\ \hline
			IoT1   & \textbf{92.093} & 56.354 \\ \hline
			IoT2   & \textbf{74.892} & 67.317 \\ \hline
			IoT3   & \textbf{95.363} & 83.647 \\ \hline
			IoT4   & \textbf{89.326} & 77.811 \\ \hline
			IoT5   & \textbf{89.006} & 82.269 \\ \hline
			IoT6   & \textbf{89.392} & 82.721 \\ \hline
			IoT7   & \textbf{98.069} & 82.397 \\ \hline
			IoT8   & \textbf{87.793} & 59.763 \\ \hline
			IoT9   & \textbf{91.417} & 66.489 \\ \hline
			KDD    & \textbf{99.576} & 84.471 \\ \hline
			NSLKDD & \textbf{98.900} & 82.406 \\ \hline
			UNSW   & \textbf{97.159} & 67.203 \\ \hline
		\end{tabular}
		\caption{The results with $p=5$. }
		\label{table:results_p5_50000}
	\end{subtable}
	\caption{The results with multiple datasets in CASE 2.}
	\label{table:results_50000}
\end{table*}

In this section, we compare the performance of FTL and the unsupervised deep learning (UDL) method in terms of the significant number of each $p$ as explained in Section~\ref{sec:Sys}. Table~\ref{table:results_10000} and Table~\ref{table:results_50000} describe the significant number of each dataset with $p=1,3,5$ corresponding to the confidence of $99\%, 97\%, 95\%$. 

In general, Table~\ref{table:results_10000} and Table~\ref{table:results_50000} show that the significant numbers of all datasets increase as $p$ increases. This is because in~(\ref{equal:sig}), we calculate the significant number based on a series of 30 continuous AUC results. When $p$ increases, the AUC results increase in all tables. This demonstrates that most of the AUC results in 30 series are higher than the significant number in the case where $p=1$. 

Table~\ref{table:results_p=5_10000} shows the significant numbers of participated datasets with $p=5$ in CASE 1. In this table, the IoT1 and UNSW datasets show a significant gap of about 30\% and 40\% between FTL and UDL. These results show the difficulty of clustering in recognizing the groups of samples and the advantage of collaborative learning in these datasets. The other ten datasets have gaps of around 10-20\% between the two methods, which demonstrate the stability of our proposed solution for any cybersecurity dataset.

In addition, Table~\ref{table:results_p5_50000} shows the significant numbers of multiple datasets with $p=5$ in CASE 2. In this table, the significant numbers also have a gap of around 10-40\% between the two solutions. It shows the common trend that the significant numbers increase for most datasets when the number of samples increases. However, in IoT2, IoT5, and IoT6 datasets, the significant numbers slightly decrease because of the randomly selected samples from the original dataset. It also can be demonstrated by the high fluctuation of the reconstruction errors of IoT2, IoT5, IoT6 datasets in Fig.~\ref{fig:loss_iot_50000} compared with other datasets. However, in all studied datasets, our proposed solution still performs much better than the state-of-the-art UDL solution. These results demonstrate that our solution can work efficiently in all IoT and conventional cybersecurity datasets in detecting cyberattacks in~the~network.  

\subsubsection{Reconstruction Error Analysis}

\begin{figure*}[ht]
	\begin{subfigure}{0.5\linewidth}
		\centering
		\includegraphics[width=\linewidth]{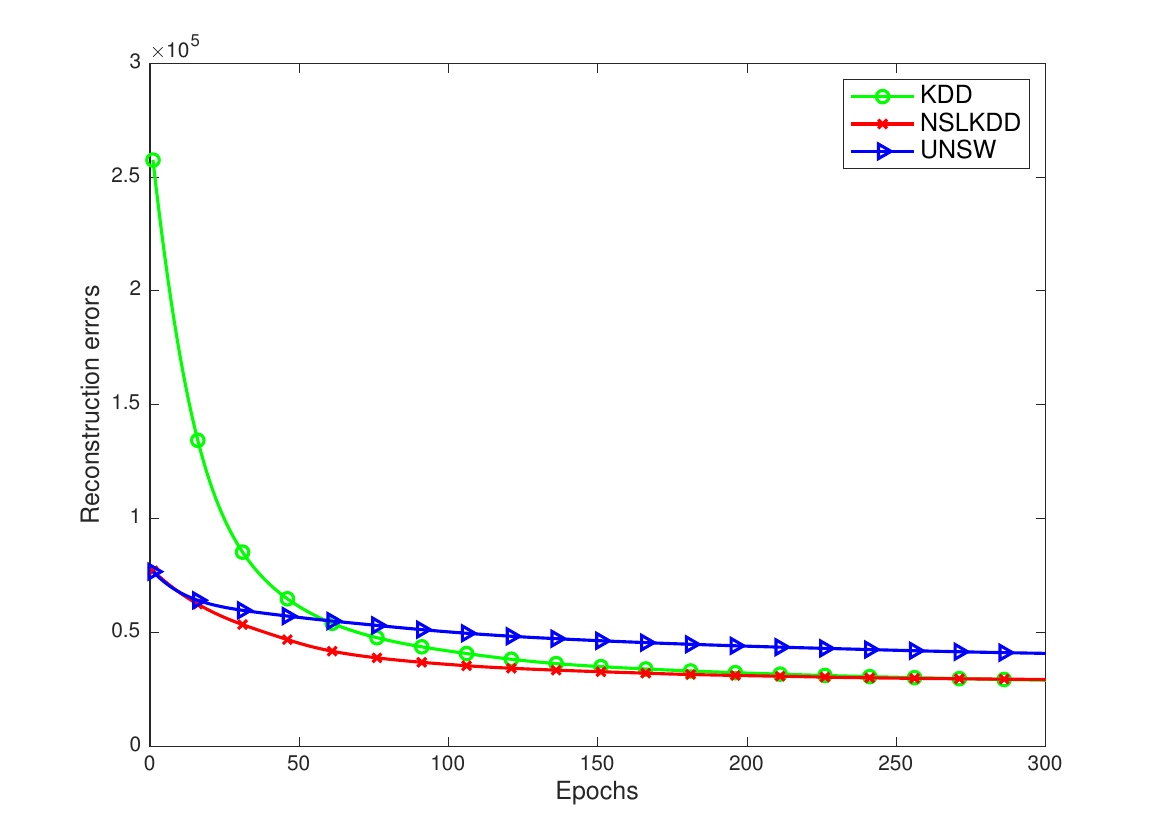}  
		\caption{The reconstruction errors of KDD, NSLKDD and UNSW datasets.}
		\label{fig:loss_traditional}
	\end{subfigure}
	\begin{subfigure}{0.5\linewidth}
		\centering
		\includegraphics[width=\linewidth]{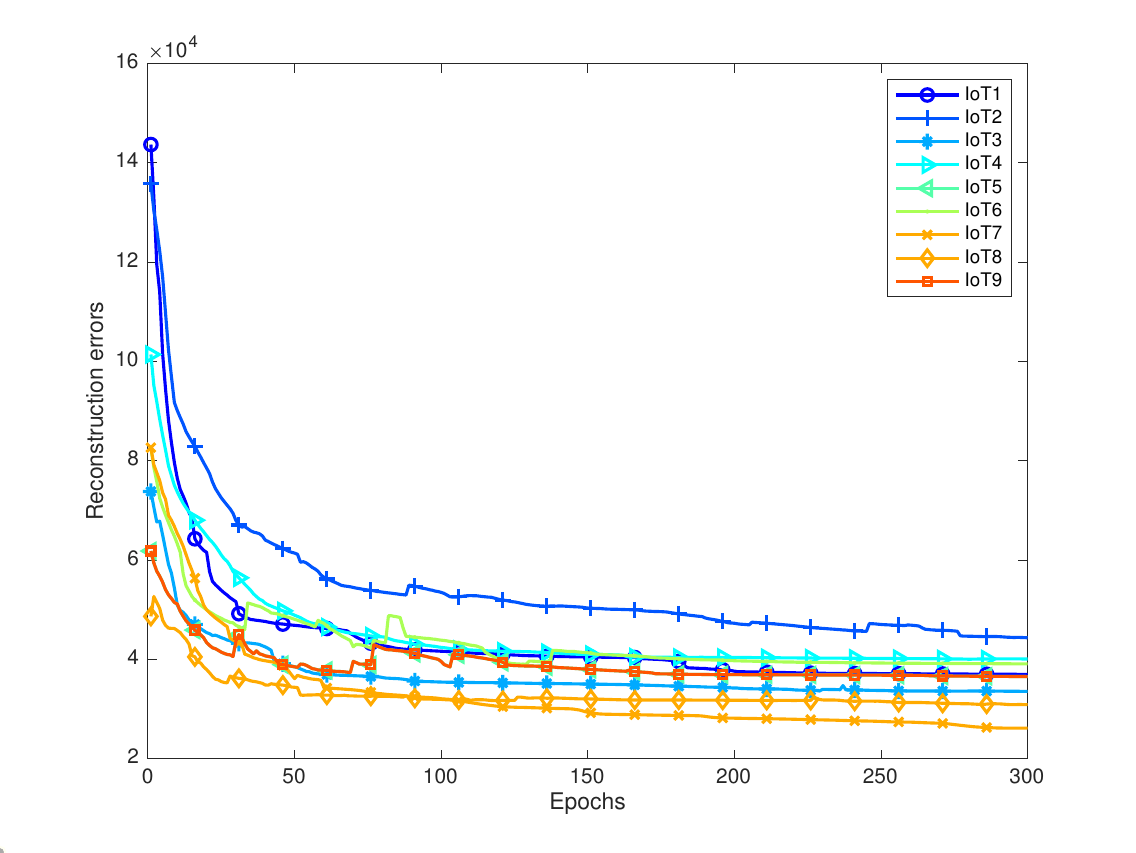}  
		\caption{The reconstruction errors of IoT datasets.}
		\label{fig:loss_iot}
	\end{subfigure}
	\caption{The reconstruction errors in CASE 1.}
	\label{fig:loss}
\end{figure*}  

\begin{figure*}[ht]
	\begin{subfigure}{0.5\linewidth}
		\centering
		\includegraphics[width=\linewidth]{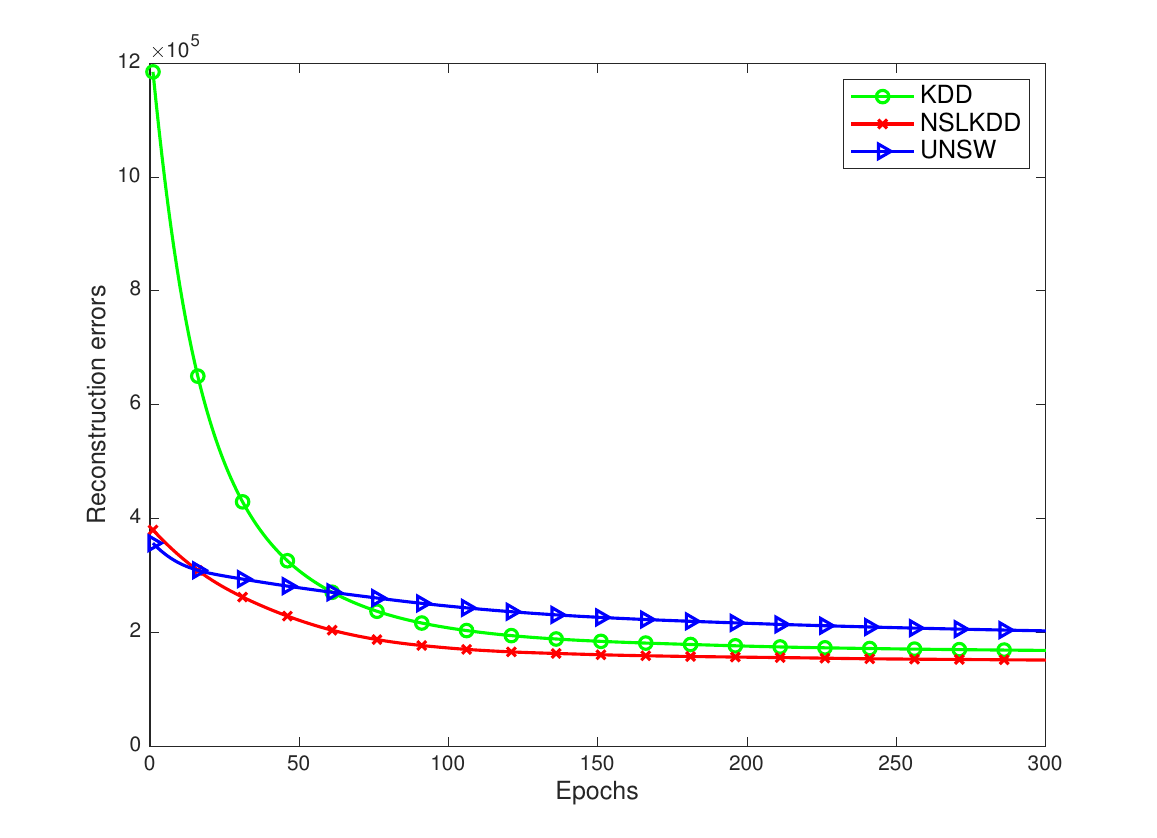}  
		\caption{The reconstruction errors of KDD, NSLKDD and UNSW datasets.}
		\label{fig:loss_traditional_50000}
	\end{subfigure}
	\begin{subfigure}{0.5\linewidth}
		\centering
		\includegraphics[width=\linewidth]{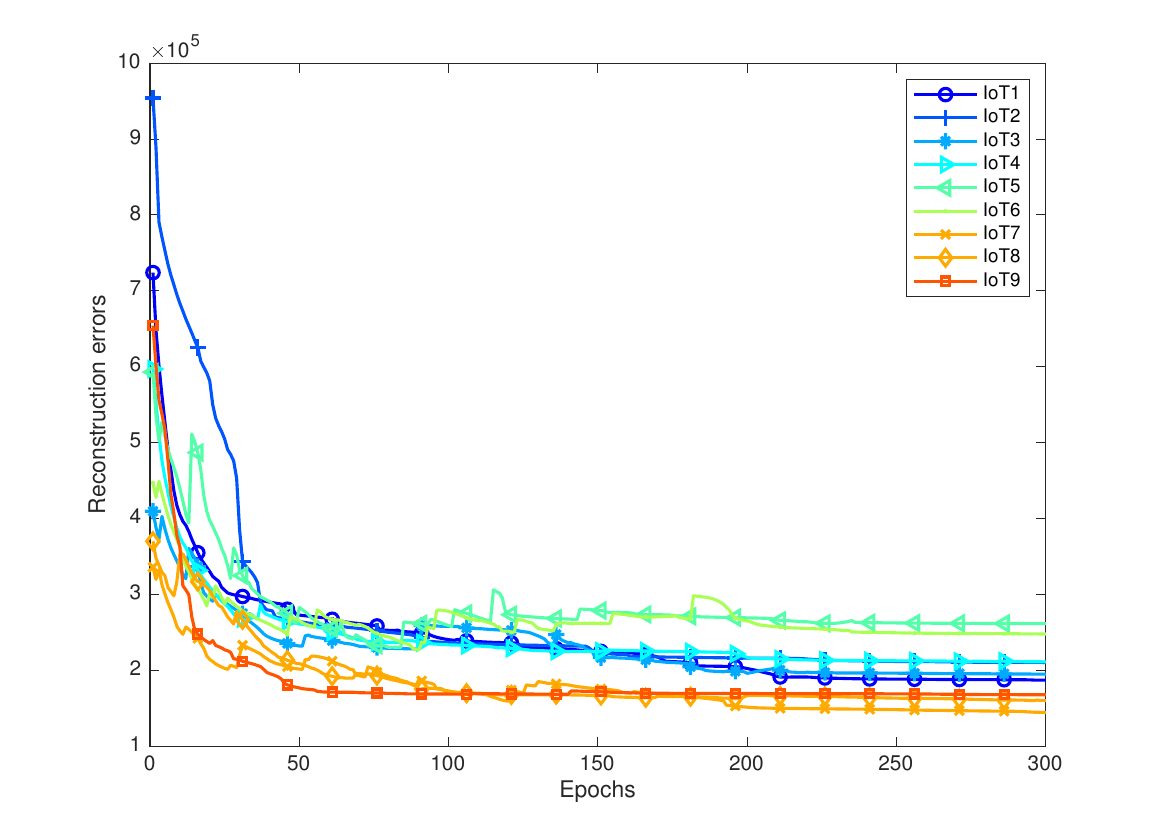}  
		\caption{The reconstruction errors of IoT datasets.}
		\label{fig:loss_iot_50000}
	\end{subfigure}
	\caption{The reconstruction errors in CASE 2.}
	\label{fig:loss_50000}
\end{figure*}

In this section, we discuss the convergence of the FTL algorithm in each dataset. Fig.~\ref{fig:loss} describes the reconstruction errors of the nine IoT datasets and the conventional datasets like KDD, NSLKDD, and UNSW in CASE 1. Fig.~\ref{fig:loss_50000} describes the reconstruction errors of study datasets in CASE~2.

In Fig.~\ref{fig:loss_traditional} and Fig.~\ref{fig:loss_traditional_50000}, we can see that at the first few epochs, the errors are very high for KDD (up to $2.6 \times 10^5$ in CASE~1 and $12 \times 10^5$ in CASE~2), but this error dramatically reduces to $0.3 \times 10^5$ in CASE~1 and $1.5 \times 10^5$ in CASE~2 after only $200$ epochs. For NSLKDD and UNSW, they have very similar trends with $0.75 \times 10^5$ in CASE 1 and $3.8 \times 10^5$ in CASE~2 at the beginning and gradually reduce to $0.4 \times 10^5$ in CASE~1 and $1.9 \times 10^5$ in CASE~2 after $200$ epochs, respectively. After 200 epochs, the algorithm converges as all the reconstruction error curves are flattened.

Fig.~\ref{fig:loss_iot} and Fig.~\ref{fig:loss_iot_50000} show the reconstruction errors of nine IoT datasets in both CASE~1 and CASE~2. we can observe the same trend over all datasets, i.e., all errors gradually reduce when the number of epochs increases. However, it can be observed that the trend exhibits some fluctuations in comparison with the trends in Fig.~\ref{fig:loss_traditional} and Fig.~\ref{fig:loss_traditional_50000} because of the heterogeneous distribution in IoT datasets. The high fluctuation of the reconstruction errors of IoT2, IoT5, IoT6 datasets in Fig.~\ref{fig:loss_iot_50000} also explains why their significant numbers reduce when the number of samples increases in CASE~2. However, the reconstruction errors of all studied datasets in our proposed solution dramatically decrease and become stable after $200$ running epochs in both cases. 

\subsubsection{Mutual Information Analysis}
\label{mutual_information}

\begin{figure*}[ht]
	\begin{subfigure}{0.5\linewidth}
		\centering
		\includegraphics[width=\linewidth]{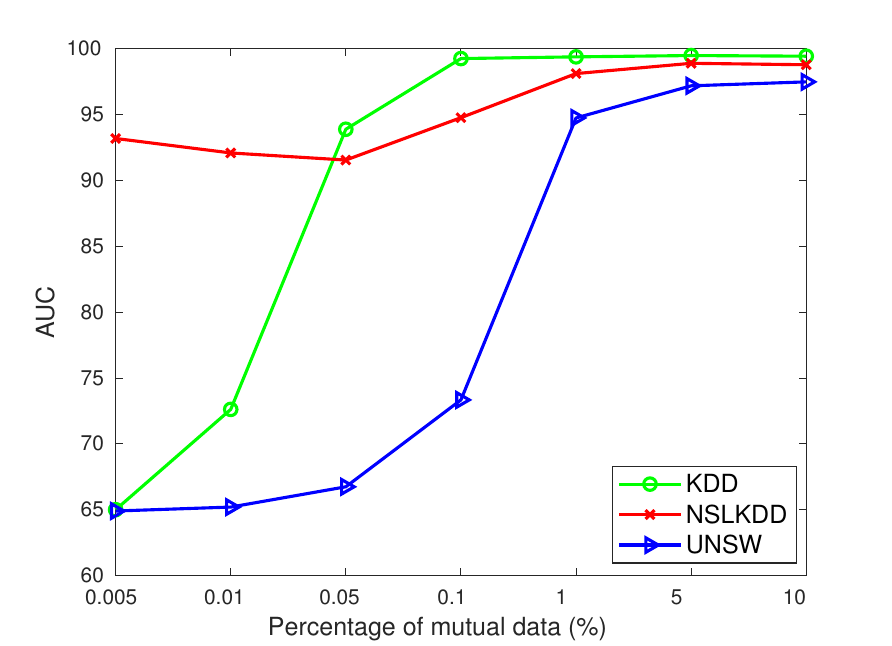}  
		\caption{The percentage mutual information of KDD, NSLKDD and UNSW datasets.}
		\label{fig:mutual_traditional}
	\end{subfigure}
	\begin{subfigure}{0.5\linewidth}
		\centering
		\includegraphics[width=\linewidth]{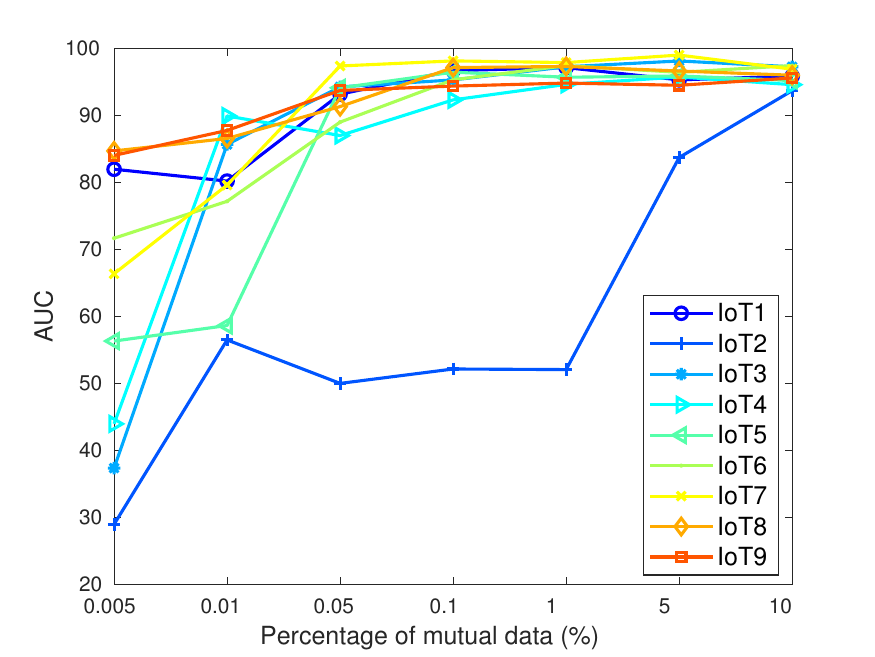}  
		\caption{The percentage mutual information of IoT datasets.}
		\label{fig:mutual_iot}
	\end{subfigure}
	\caption{The illustration of AUC with different percentage of mutual information.}
	\label{fig:mutual}
\end{figure*}

As mentioned in the previous section, network A and network B may share a number of mutual samples. The FTL algorithm exploits the information of these mutual samples to perform the prediction for unlabeled data of network B. This section provides the analysis results to identify how this mutual information can affect to the results of label prediction. In this section, we perform the simulation in CASE 2 with a larger number of samples than in CASE 1. Fig.~\ref{fig:mutual} gives information about the variation of AUC when the percentage of mutual data~increases. 

Fig.~\ref{fig:mutual_traditional} shows the increase of AUC on KDD, NSLKDD, and UNSW datasets when the percentage of mutual samples increases from 0.005\% to 10\%. The AUC of KDD and UNSW datasets sharply increase and remain stable at around 96\% on the NSLKDD dataset with about 5\% to 10\% mutual samples. A similar trend happens with the IoT datasets in Fig.~\ref{fig:mutual_iot} when the AUCs of all nine IoT datasets increase and remain stable at approximately 10\% of mutual samples. From these results, it can be observed that achieving high efficiency in AUC for IoT datasets may require at least 10\% of mutual data. 

In summary, the results with 12 cybersecurity datasets show the outperformance of our proposed model in comparison with the state-of-the-art unsupervised deep learning in term of accuracy as shown in Table~\ref{table:results_10000} for CASE 1 and Table~\ref{table:results_50000} for CASE 2, especially with IoT1 and UNSW datasets. Moreover, the reconstruction errors show a fluctuation of the IoT datasets when the number of samples increases due to noise from the collected datasets of some IoT devices. Finally, we vary the amount of mutual data between two networks to evaluate the accuracy of our proposed model. The results show that the proposed model can achieve high performance with 10\% mutual data with all datasets.

\section{Conclusion}
\label{sec:Conc}

In this work, we have proposed a novel collaborative learning framework to address the limitations of current ML-based cyberattack detection systems in IoT networks. In particular, by extracting and transferring knowledge from a network with abundant labeled data (source network), the intrusion detection performance of the target network can be significantly improved (even if the target has very few labeled data). More importantly, unlike most of the current works in this area, our proposed framework can enable the source network to transfer the knowledge to the target network even when they are different data structures, e.g., different features. The experimental results then show that the accuracy of prediction of our proposed framework is significantly improved in comparison with the state-of-the-art unsupervised deep learning model. In addition, the convergence of the proposed collaborative learning model is also analyzed with various cybersecurity datasets. In future work, we can consider using other effective transfer learning techniques to make transfer learning processes more stable and achieve better performance, especially when the amount of mutual information is very~limited.  

\section{Acknowledgements}
\label{sec:Ack}

This work is the output of the ASEAN IVO \url{http://www.nict.go.jp/en/asean_ivo/index.html} project ``Cyber-Attack Detection and Information Security for Industry 4.0'' and financially supported by NICT \url{http://www.nict.go.jp/en/index.html}.

This work was supported in part by the Joint Technology and Innovation Research Centre -- a partnership between the University of Technology Sydney and the University of Engineering and Technology, Vietnam National University, Hanoi, Vietnam.

This research was supported in part by the Australian Research Council under the DECRA project DE210100651.

The work of Cong T. Nguyen was funded in part by Vingroup JSC and supported in part by the Master, PhD Scholarship Programme of Vingroup Innovation Foundation (VINIF), Institute of Big Data, code VINIF.2021.TS.006.

\bibliographystyle{IEEEtran}
\bibliography{reference2}

\newpage

\vfill

\end{document}